\documentclass[peerreviw]{IEEEtran}

\IEEEoverridecommandlockouts

\usepackage{amsmath,amssymb,amsfonts}
\usepackage{algorithmic}
\usepackage{graphicx}
\usepackage{textcomp}
\usepackage{xcolor}
\usepackage[hidelinks]{hyperref}
\usepackage[justification=centering]{caption}
\usepackage{svg}
\usepackage{multirow}
\usepackage{makecell}
\usepackage{booktabs}
\usepackage[acronym]{glossaries}
\usepackage{pifont}
\usepackage{float}

\usepackage[maxbibnames=4,doi=false,style=ieee,backend=biber]{biblatex}
\usepackage[capitalise]{cleveref}
\def\BibTeX{{\rm B\kern-.05em{\sc i\kern-.025em b}\kern-.08em
    T\kern-.1667em\lower.7ex\hbox{E}\kern-.125emX}}

\begin{document}
\newacronym{vc}{VC}{Voice Cloning}
\newacronym{tts}{TTS}{Text-to-Speech}
\newacronym{for}{FoR}{Fake or Real}
\newacronym{gans}{GANs}{Generative Adversarial Networks}
\newacronym{mfccs}{MFCCs}{Mel-Frequency Cepstrum Coefficients}
\newacronym{lfccs}{LFCCs}{Linear Frequency Cepstral Coefficients}
\newacronym{gmm}{GMM}{Gaussian Mixture Model}
\newacronym{stft}{STFT}{Short Time Fourier Transform}
\newacronym{eer}{EER}{Equal Error Rate}

\newcommand{\etal}{\textit{et al}.\@ }
\newcommand{\ie}{\textit{i.e.},\@ }
\newcommand{\eg}{\textit{e.g.},\@ }

\title{HISPASpoof: A New Dataset For\\ Spanish Speech Forensics}

\newcommand{\td}{$^\dagger$}
\newcommand{\tdd}{$^\ddagger$}

\author{
\parbox{0.95\linewidth}{
\hspace*{\fill} Maria Risques, Kratika Bhagtani, Amit Kumar Singh Yadav and Edward J. Delp \hspace*{\fill}
    \vspace*{0.1em}\\
\small\centering Video and Image Processing Laboratory (VIPER), School of Electrical and Computer Engineering,\\ Purdue University. West Lafayette, Indiana, USA
}
}
\vspace{0.1cm}
\maketitle
\thispagestyle{empty}
\pagestyle{empty}

\begin{abstract}
Zero-shot Voice Cloning (VC) and Text-to-Speech (TTS) methods have advanced rapidly, enabling the generation of highly realistic synthetic speech and raising serious concerns about their misuse. While numerous detectors have been developed for English and Chinese, Spanish—spoken by over 600 million people worldwide—remains underrepresented in speech forensics. To address this gap, we introduce HISPASpoof, the first large-scale Spanish dataset designed for synthetic speech detection and attribution. It includes real speech from public corpora across six accents and synthetic speech generated with six zero-shot TTS systems. We evaluate five representative methods, showing that detectors trained on English fail to generalize to Spanish, while training on HISPASpoof substantially improves detection. 
We also evaluate synthetic speech attribution performance on HISPASpoof, i.e., identifying the generation method of synthetic speech.
HISPASpoof thus provides a critical benchmark for advancing reliable and inclusive speech forensics in Spanish.

\end{abstract}


\glsresetall

\section{Introduction}\label{sec:intro}

The rapid advancement of speech synthesis techniques has significantly transformed the area of audio generation and speech forensics. 
Recent Text-to-Speech (TTS) and Voice Cloning (VC) methods~\cite{elevenlabs, f5spanish, fishspeech2024arxiv, xttsv1, xttsv2, yourtts2021} are now capable of producing highly realistic synthetic voices that closely mimic the spectral, prosodic, and linguistic traits of real human speech~\cite{bank,couple,scammers, loved}. 
Zero-shot VC methods can generate synthetic speech of a target speaker using only a few seconds of their real speech as reference~\cite{arik2018,singh2024}. 
These technologies have applications in domains such as virtual assistants~\cite{siri}, and media production~\cite{films2022}. Their increasing availability has heightened concerns regarding their malicious use for misinformation~\cite{zelensky2022}, impersonation~\cite{david2018}, and fraud~\cite{smith2021}. This emphasizes the need for reliable detection and attribution methods.
Synthetic speech detection aims to distinguish between real and generated speech, whereas attribution goes further by identifying the synthesizer that was used to generate synthetic speech.

Despite the development of various synthetic speech detectors~\cite{ps3dt, tak22_odyssey, koutini2022passt, tssdnet},
most efforts in speech forensics have been concentrated on English and Chinese speech with limited exploration of other languages. 
Popular speech forensics datasets such as ASVspoof2019~\cite{asvdata_2019}, ASVspoof2021~\cite{asvspoof_2021}, Fake or Real (FoR)~\cite{reimao2019}, TIMIT-TTS~\cite{salvi2023}, In-the-Wild~\cite{müller2024},  and Chinese Dataset for Fake Audio Detection (CFAD)~\cite{MA2024103122}, focus exclusively on English or Chinese speech. 
Synthetic speech detection and attribution methods trained on these datasets~\cite{lfcc, ps3dt, tak22_odyssey, koutini2022passt, tssdnet,alzantot2019, fgssat2023, amit_ei_paper} are often optimized for the phonetic and prosodic characteristics of these languages. 
As a result, there exists a  gap in the applicability of these methods to other languages—particularly Spanish.

Spanish is one of the most widely spoken languages in the world, with more than 600 million speakers~\cite{cervantes_2024}, equivalent to 7.5\% of the global population~\cite{cervantes_2024}. 
Nearly 500 million people speak Spanish as their native language~\cite{cervantes_2024}, making it the second most spoken mother tongue globally, after Chinese. Spanish is the official language in 20 countries and several territories, primarily across Latin America, as well as in Spain and Equatorial Guinea~\cite{cervantes_2024}. Beyond these regions, Spanish maintains a significant presence in non-Hispanic countries, most notably in the United States~\cite{cervantes_2024}. The United States is home to more than 60 million Hispanic residents~\cite{cervantes_2024}, of whom approximately 75\% are proficient in Spanish~\cite{cervantes_2024}, making it the country with the second largest population of Spanish speakers in the world~\cite{cervantes_2024}. 
This highlights the global importance of Spanish not only as a native language but also as a language of international communication, education, and media~\cite{cervantes_2024}.

The lack of large Spanish datasets limits the development of detection and attribution methods for Spanish speech.
A few multilingual corpora include Spanish speech, such as ODSS (Open Dataset of Synthetic Speech)~\cite{yaroshchuk2023}. 
However, the quantity of Spanish data and generator diversity remain insufficient for comprehensive training and evaluation.
To the best of our knowledge, HISPASpoof fills this gap as the first large-scale Spanish dataset supporting both detection and attribution. 

In this paper, we show that synthetic speech detectors~\cite{lfcc, alzantot2019, koutini2022passt, tak22_odyssey} that perform well in English, fail to generalize to Spanish when trained solely on English speech. We evaluate five representative methods: Linear Frequency Cepstral Coefficients with Gaussian Mixture Model (LFCC-GMM)~\cite{lfcc}, Mel-Frequency Cepstral Coefficients with Residual Network (MFCC-ResNet)~\cite{alzantot2019}, Spectrogram with Residual Network (Spec-ResNet)~\cite{alzantot2019}, Patchout faSt Spectrogram
Transformer (PaSST)~\cite{koutini2022passt}, and Wav2Vec2-based Audio Anti-Spoofing using Integrated Spectro-Temporal network (AASIST)~\cite{tak22_odyssey}.
We propose HISPASpoof, a dataset of real and synthetic Spanish speech. 
It contains two complementary subsets: a detection subset and an attribution subset. 
Real Spanish speech signals are taken from publicly available corpora that include six Spanish accents: Colombian~\cite{guevara2020crowdsourcing}, Argentinian~\cite{guevara2020crowdsourcing}, Chilean~\cite{guevara2020crowdsourcing}, Mexican~\cite{hernandezmena2014ciempiess}, Peruvian~\cite{guevara2020crowdsourcing} and Peninsular~\cite{wang2021voxpopuli}, ensuring broad phonetic coverage. We generate synthetic speech using six zero-shot TTS methods: XTTS-v1~\cite{xttsv1}, XTTS-v2~\cite{xttsv2}, YourTTS~\cite{yourtts2021}, FishSpeech~\cite{fishspeech2024arxiv}, ElevenLabs~\cite{elevenlabs}, and F5-TTS~\cite{f5spanish}. 
These methods allow speech synthesis for any target speaker using only a short reference speech, without requiring speaker-specific training.

We demonstrate that training on the HISPASpoof dataset substantially improves detection performance on Spanish speech. 

\glsresetall
\section{Speech Forensics Background}\label{sec:related_works}

Several public challenges and datasets, notably ASVspoof2019~\cite{asvdata_2019} and ASVspoof2021~\cite{asvspoof_2021}, have advanced speech forensics research.
In addition, several other datasets have been proposed, including Fake or Real (FoR)~\cite{reimao2019}, TIMIT-TTS~\cite{salvi2023}, and In-the-Wild~\cite{müller2024}. These datasets have enabled standardized evaluation of synthetic speech detection and attribution methods, but they are limited to English. Other datasets, such as CFAD~\cite{MA2024103122}, have been developed specifically for Chinese.
Among the few multilingual datasets available, the ODSS dataset~\cite{yaroshchuk2023} stands out as it includes Spanish, English and German speech. 
However, it contains fewer than 7,000 Spanish speech signals, and the synthetic speech is generated using only two voice cloning techniques, which may not reflect the full spectrum of modern TTS systems. 

Several synthetic speech detection and attribution methods have been proposed, which can be categorized into three main categories based on how they process input speech~\cite{bhagtani2022overviewrecentworkmedia,fairSSD1}: feature-based, image-based, and waveform-based.

\textbf{Feature-based} approaches extract acoustic descriptors such as Mel-Frequency Cepstral Coefficients (MFCCs), Linear Frequency Cepstral Coefficients (LFCCs), and Constant-Q Cepstral Coefficients (CQCCs), which are sensitive to spectral anomalies in synthetic signals~\cite{li2021replay, akdeniz2021detection, assd_2023}. These features are typically processed using classifiers such as Gaussian Mixture Models (GMMs), Convolutional Neural Networks (CNNs), or recurrent architectures such as Long Short-Term Memory (LSTM) networks.

\textbf{Image-based} methods convert speech signals into two-dimensional time-frequency representations (e.g., spectrograms or mel-spectrograms), allowing synthetic speech forensics methods to treat the input as an image and apply visual feature extraction techniques~\cite{bartusiak2023, yadav2023dsvae}. This approach facilitates the use of vision-based deep learning methods, such as ResNet~\cite{alzantot2019} and transformers~\cite{bartusiak2023,koutini2022passt, mdrt_amit, acm_kratika_2023}, to learn discriminative patterns.

\textbf{Waveform-based} techniques bypass the need for handcrafted features or transformations, operating directly on raw speech waveforms. Leveraging methods such as Wav2Vec2~\cite{improvingfair}, convolutional-recurrent hybrids, or attention-based networks, these methods aim to capture the fine-grained temporal dynamics of speech signals~\cite{tak2022odyssey}.
\glsresetall
\section{Experimental Datasets}\label{sec:sssd_dataset}

In this section, we first describe the two datasets used in our study: the ASVspoof2019 dataset~\cite{asvdata_2019} (English) and the ODSS dataset~\cite{yaroshchuk2023} (English, German, and Spanish).
Next, we describe the development of our proposed dataset, HISPASpoof (Spanish).

\subsection{The ASVspoof2019 Dataset:} This dataset~\cite{asvdata_2019} is commonly used for synthetic speech detection~\cite{ps3dt, tak22_odyssey, koutini2022passt, tssdnet,alzantot2019} and attribution~\cite{fgssat2023}.
It contains real and synthetic speech in English. 
In our experiments, we specifically use the Logical Access (LA) subset of the dataset. 
The synthetic speech is generated using 17 different synthesizers (A01-A19). 
Most synthesizers appear only in the training/development sets or only in the evaluation set, except for synthesizers A04 and A06 which appear in both training and evaluation.
Training, development, and evaluation sets are speaker-disjoint.


\subsection{The ODSS Dataset:} This dataset~\cite{yaroshchuk2023} covers Spanish, English, and German speech. The dataset includes 6,771 Spanish signals from 96 speakers across five accents, 17,476 English signals from 56 speakers, and 5,778 German signals from 4 speakers, encompassing both real and synthetic speech signals.
There are in total, 30,025 speech signals and only two speech synthesizers are included. 
With disjoint constraints for unseen speakers and generators, the effective Spanish training set is reduced to just 3,618 signals.

\subsection{The HISPASpoof Dataset:} 
We introduce a new dataset for speech forensics in Spanish, developed using real speech from publicly available corpora~\cite{guevara2020crowdsourcing, hernandezmena2014ciempiess, wang2021voxpopuli}. 
HISPASpoof contains two complementary subsets: a \emph{detection subset} for distinguishing real from synthetic speech, and an \emph{attribution subset} for identifying the synthesizer that generated a given synthetic speech. 

Both subsets share the same design principles: gender-balanced speakers across six Spanish accents (Colombian, Argentinian, Chilean, Mexican, Peruvian, and Peninsular) and the use of six recent zero-shot TTS systems (XTTS-v1~\cite{xttsv2}, XTTS-v2~\cite{xttsv2}, YourTTS~\cite{yourtts_github}, FishSpeech~\cite{fishspeech2024arxiv}, ElevenLabs~\cite{elevenlabs}, and F5-TTS~\cite{f5spanish}). The dataset includes 4 speakers per accent.
The dataset is divided into training, validation, and test splits, with the test set containing both unseen speakers and unseen generators. 
This setup enables realistic evaluation of generalization capabilities in Spanish-language synthetic speech detection and attribution.

\textbf{The Detection Subset} contains both synthetic and and real speech. Table~\ref{tab:real_speech_distribution} summarizes the six Spanish accents included in the dataset, along with their respective licenses, sampling rates ($\textbf{S}_\textbf{r}$) and the number of real speech signals in this subset. Synthetic speech was generated using the exact transcript of each real signal to avoid bias with respect to speech content. We use six zero-shot TTS systems for synthesis, as summarized in Table~\ref{tab:dataset_distribution_combined}. Overall, the HISPASpoof detection subset contains 43,687 Spanish speech signals, out of which 37,446 are synthetic and 6,241 are real.

\begin{table}[ht]
\vspace{-0.4cm}
    \centering
    \renewcommand{\arraystretch}{1.2}
    \caption{Distribution of real speech across different Spanish accents.}
    \label{tab:real_speech_distribution}
    \begin{tabular}{l@{\hspace{-0.6cm}} c c@{\hspace{0.1cm}} c @{\hspace{0.2cm}} r}
        \toprule
        \textbf{Source} & & \textbf{License} & $\textbf{S}_\textbf{r}$ (\textbf{kHz}) & \textbf{Signals} \\
        \midrule
        A1 & Peninsular Spanish \cite{ wang2021voxpopuli} & CC0 & 16 & \textbf{2,135} \\
        A2 & Argentinian Spanish \cite{guevara2020crowdsourcing} & CC BY-SA 4.0 & 48 & \textbf{522} \\
        A3 & Colombian Spanish \cite{guevara2020crowdsourcing} & CC BY-SA 4.0 & 48 & \textbf{595} \\
        A4 & Mexican Spanish \cite{hernandezmena2014ciempiess} & CC BY-SA 4.0 & 16 & \textbf{1,892} \\
        A5 & Chilean Spanish \cite{guevara2020crowdsourcing} & CC BY-SA 4.0 & 48 & \textbf{510} \\
        A6 & Peruvian Spanish \cite{guevara2020crowdsourcing} & CC BY-SA 4.0 & 48 & \textbf{587} \\
        \midrule
        \textbf{Total Real} & & -- & -- & \textbf{6,241} \\
        \bottomrule
    \end{tabular}
\end{table}

\textbf{The Attribution Subset} was generated using 4,000 Spanish text lines generated with ChatGPT-4o, a Large Language Model (LLM) developed by OpenAI~\cite{cite44, cite45}. 
ChatGPT was selected to produce copyright-free material while ensuring diversity across domains such as history, conversations, sports, and weather. 
The generated text was processed to eliminate redundancy. 
The resulting text corpus contains lines ranging from 4 to 46 words, with an average of 18 words per line, each consisting of one or more sentences. 
The same 4,000 lines were synthesized for each speaker across all five synthesizers to avoid content bias, resulting in 96,000 samples per synthesizer.
For ElevenLabs, due to usage costs, only 500 lines out of the total 4000 lines per speaker were synthesized, yielding 12,000 samples. 
The complete distribution of synthetic speech is shown in Table~\ref{tab:dataset_distribution_combined}. 

\begin{table}[b!]
    \centering
    \renewcommand{\arraystretch}{1.2}
    \caption{Distribution of real and synthetic Spanish speech in the HISPASpoof dataset. CM and OS denote commercial and open-source synthesizers, respectively.}
    \label{tab:dataset_distribution_combined}
    \setlength{\tabcolsep}{9.5pt}
    \begin{tabular}{l c c c c}
        \toprule
        \multicolumn{2}{l}{\textbf{Subset / Source}} & \textbf{License} & $\mathbf{S}_\mathbf{r}$\textbf{(kHz)} & \textbf{Samples} \\
        \midrule
        \multicolumn{5}{l}{\textbf{Detection Subset}} \\
        \midrule
        R & Real Speech & -- & 16/48 & \textbf{6,241} \\
        G1 & ElevenLabs~\cite{elevenlabs} & CM & 44.1 & \textbf{6,241} \\
        G2 & F5-Spanish~\cite{f5spanish} & OS & 24 & \textbf{6,241} \\
        G3 & FishSpeech~\cite{fishspeech2024arxiv} & OS & 44.1 & \textbf{6,241} \\
        G4 & XTTS-v1.1~\cite{xttsv1} & OS & 24 & \textbf{6,241} \\
        G5 & XTTS-v2~\cite{xttsv2} & OS & 22.05 & \textbf{6,241} \\
        G6 & YourTTS~\cite{yourtts_github} & OS & 16 & \textbf{6,241} \\
        \multicolumn{2}{l}{\textbf{Total Detection}} & -- & -- & \textbf{43,687} \\
        \midrule
        \multicolumn{5}{l}{\textbf{Attribution Subset}} \\
        \midrule
        G1 & ElevenLabs~\cite{elevenlabs} & CM & 44.1 & \textbf{12,000} \\
        G2 & F5-Spanish~\cite{f5spanish} & OS & 24 & \textbf{96,000} \\
        G3 & FishSpeech~\cite{fishspeech2024arxiv} & OS & 44.1 & \textbf{96,000} \\
        G4 & XTTS-v1.1~\cite{xttsv1} & OS & 24 & \textbf{96,000} \\
        G5 & XTTS-v2~\cite{xttsv2} & OS & 22.05 & \textbf{96,000} \\
        G6 & YourTTS~\cite{yourtts_github} & OS & 16 & \textbf{96,000} \\
        \multicolumn{2}{l}{\textbf{Total Attribution}} & -- & -- & \textbf{492,000} \\
        \midrule
        \textbf{Total} & & -- & -- & \textbf{535,687} \\
        \bottomrule
    \end{tabular}
\end{table}

In both subsets, we partition speakers into two distinct groups to enable evaluation of both within-distribution and out-of-distribution generalization performance:

\begin{itemize} 
    \item \textbf{Seen Speakers:} Speech from these speakers is used for training and validation. A portion is also included in the test set to assess within-distribution generalization.
    \item \textbf{Unseen Speakers:} Reserved exclusively for the test set to evaluate performance on unseen voices. 
\end{itemize} 

To construct the unseen speaker group, we select one speaker per Spanish accent (either male or female), ensuring diversity in both linguistic and demographic characteristics. The remaining speakers are assigned to the seen speaker group, with their signals partitioned into 82.5\% for training, 5\% for validation, and 12.5\% for testing.
Additionally, to assess generalization to unseen synthesis techniques, two VC methods (XTTSv1 and Fish-Speech) are reserved exclusively for testing. These generators are not used during training or validation, providing a challenging and realistic evaluation scenario.
The test set includes both - unseen speakers and synthesizers
(we refer to this as the Unseen Subset ($U$)), as well as the signals from speakers and synthesizers encountered during training (we refer to this as the Seen Subset ($S$)). 
For the remaining speech signals, the distribution in the \emph{detection subset} is 
50\% for training, 5\% for validation, and 45\% for testing. 
For the \emph{attribution subset}, the distribution is 
35\% for training, 10\% for validation, and 55\% for testing. Overall, our setup enables the evaluation of detection methods under both closed-set (Seen Subset, $S$) and open-set (Unseen Subset, $U$) conditions. 
The overall distribution of speech across training, validation, and test sets is shown in Table~\ref{tab:dataset_distribution3}.

\begin{table}[t!]
\vspace{-0.4cm}
    \centering
    \caption{Dataset distribution: training, validation and test sets}
    \label{tab:dataset_distribution3}
    \renewcommand{\arraystretch}{1.2}
    \begin{tabular}{l c c c r}
        \toprule
        \textbf{Source} & \textbf{Train} & \textbf{Validation} & \textbf{Test} & \textbf{Total} \\
        \midrule
        \textbf{Detection Subset} & 20,945 & 2,619 &  20,123 & 43,687 \\
         \textbf{Attribution Subset} & 168,750 & 45,000 & 278,250  &  492,000 \\
        \bottomrule
    \end{tabular}
\vspace{-0.2cm}
    
\end{table}


We make the HISPASpoof dataset publicly available~\cite{gits3d}: https://gitlab.com/viper-purdue/s3d-spanish-syn-speech-det.git

\glsresetall
\section{Detection and Attribution Methods}\label{sec:detection-methods}

We evaluate five methods representative
of the three main categories described in Section~\ref{sec:related_works},~\ie feature-based, image-based, and waveform-based approaches that have shown strong performance on English datasets~\cite{lfcc, alzantot2019, koutini2022passt, tak22_odyssey}.
Four of these (MFCC-ResNet, Spec-ResNet, PaSST, Wav2Vec2-AASIST) are evaluated for both detection and attribution tasks, while LFCC-GMM is used only for detection as it has been primarily developed and evaluated for binary classification tasks in prior work.

Table~\ref{tab:detection_methods_overview} summarizes the type of input feature used for each method, type of network, and the number of trainable parameters. The five speech forensics methods used in this study are implemented based on the code provided in~\cite{gits3d}: https://gitlab.com/viper-purdue/s3d-spanish-syn-speech-det.git

\vspace{0.2cm}

\begin{table}[H]
\caption{Detection Methods Overview}
    \label{tab:detection_methods_overview}
    \centering
    \renewcommand{\arraystretch}{1.2}
    \begin{tabular}{l@{\hspace{0.2cm}}c@{\hspace{0.2cm}}c@{\hspace{0.2cm}}r}
        \toprule
        \textbf{Method} & \textbf{Input Feature} & \textbf{Network} & \textbf{Parameters} \\
        \midrule
        \textbf{LFCC+GMM}~\cite{lfcc} & Hand-crafted & GMM & 0.1M\\
        \textbf{MFCC-ResNet}~\cite{alzantot2019} & Hand-crafted & ResNet & 0.26M \\
        \textbf{Spec-ResNet}~\cite{alzantot2019} & Log-Spect & ResNet & 0.32M \\
        \textbf{PaSST}~\cite{koutini2022passt} & Mel-Spect & Transformer & 85M \\
        \textbf{Wav2Vec2-AASIST}~\cite{tak22_odyssey} & Time domain & Transformer & 317M\\
        \bottomrule
    \end{tabular}
    \vspace{-0.2cm}
\end{table}

\textbf{LFCC+GMM (M1)}~\cite{lfcc} : This classical approach models real and synthetic speech distributions using Gaussian Mixture Models (GMMs) trained on Log-Frequency Cepstral Coefficients (LFCCs). We compute features using 30ms analysis windows with 15ms overlap, and retain frequency components up to 4 kHz. A separate GMM with 512 components is trained per class (real or synthetic), and test signals are classified as synthetic or real based on maximum likelihood estimation.

\vspace{0.1cm}
\textbf{MFCC+ResNet (M2)}~\cite{alzantot2019}: This method combines Mel-Frequency Cepstral Coefficients (MFCCs) with a ResNet architecture. 24 MFCCs coefficients are extracted with their first and second derivatives, producing 72-dimensional feature vectors. These are input to a ResNet trained from scratch.

\vspace{0.1cm}
\textbf{Spec+ResNet (M3)}~\cite{alzantot2019}: We use this method as a representative of the image-based category, specifically those that process spectrogram images using neural networks. 
We trained this method from scratch.
This method obtains the spectrogram by taking STFT of the input speech and squaring its magnitude. The scale is changed to a logarithmic scale. A ResNet network  processes the logarithmic-scaled spectrogram for synthetic speech detection and attribution.

\vspace{0.1cm}
\textbf{PaSST (M4)}~\cite{koutini2022passt}: The Patchout faSt Spectrogram Transformer (PaSST) is a transformer-based method that processes mel-spectrograms. This architecture uses self-attention to model long-range dependencies across both time and frequency. We fine-tune the pretrained version of this method.

\vspace{0.1cm}
\textbf{Wav2Vec2-AASIST (M5)}~\cite{tak22_odyssey}: This approach represents waveform-based methods and uses a large self-supervised neural network, Wav2Vec2, to process raw speech waveforms. We fine-tune the XLS-R version with pretrained weights provided by the authors.

\glsresetall
\section{Detection Experiments}\label{sec:experiments}

This section presents detection experiments with five detectors on the datasets described in Section~\ref{sec:sssd_dataset}.

We adopt Equal Error Rate (EER) as our primary evaluation metric, consistent with ASVspoof challenge standards and facilitating comparison with existing work.
EER corresponds to the operating point on the ROC curve where the False Acceptance Rate (FAR) equals the False Rejection Rate (FRR). A lower ($\downarrow$) EER indicates better detection performance. 
\vspace{-0.1cm}
\begin{equation}
\text{EER} = \text{FAR}(\theta^*) = \text{FRR}(\theta^*)
\end{equation}

\noindent where the optimal threshold $\theta^*$ is defined as:
\begin{equation}
\theta^* = \arg\min_{\theta} |\text{FAR}(\theta) - \text{FRR}(\theta)|
\end{equation}

\noindent FAR and FRR at a given threshold $\theta$ are computed as:

\begin{equation}
\text{FAR}(\theta) = \frac{\text{Synthetic speech classified as real}}{\text{Total synthetic speech}}
\end{equation}
\vspace{-0.2cm}
\begin{equation}
\text{FRR}(\theta) = \frac{\text{Real speech classified as synthetic}}{\text{Total real speech}}
\end{equation}

\vspace{0.3cm}
Detection methods are trained on different datasets, followed by cross-evaluation, to analyze their performance.

The test sets used for the detection experiments consist exclusively of the subsets with speakers and generators unseen during training, to ensure consistency. $U_{ASV}$, $U_{ODSS}$, $U_{ODSS_{Spa}}$, $U_{HIS}$ refer to the Unseen Subset of ASVspoof2019, ODSS, Spanish subset of ODSS and HISPASpoof, respectively. The detection experiments are:


\vspace{0.5cm}

\textbf{Experiment 1 - Training on ASVspoof2019 Dataset (English):}  
    All methods are first trained using the ASVspoof2019 dataset. 
    Then, they are evaluated on all test sets to assess both baseline performance in English and whether the methods generalize to Spanish speech.

\textbf{Experiment 2 - Training on ODSS Dataset (Multilingual):}  
    We train on the ODSS dataset using all three languages (German, English, Spanish) due to limited Spanish speech in the dataset. Methods are evaluated on all test sets to assess generalization to Spanish speech.

\textbf{Experiment 3 - Training on the Spanish Subset of ODSS (Spanish):} To further analyze cross-lingual generalization and isolate the performance on Spanish speech, detection methods are trained only on the Spanish subset of ODSS and evaluated on all test sets.

\textbf{Experiment 4 - Training on the HISPASpoof Dataset (Spanish):}  
    Finally, methods are trained on HISPASpoof dataset and evaluated on all test sets.

\section{Attribution Experiments}\label{sec:attribution-experiments}

This section presents attribution experiments on the HISPASpoof attribution subset using closed-set and open-set scenarios.
In attribution, the goal is to predict the generator $y \in \{0, 1, \dots, N\}$ of a given speech signal $s$, where $y=0,\dots,N-1$ correspond to the synthesizers used in training. For open-set evaluation, we additionally define a label $y=N$ for speech generated by synthesizers not seen during training.

We use the following evaluation metrics for attribution:

\begin{itemize}
    \item \textbf{Accuracy (Acc)} measures the proportion of correct predictions:
    \[
    \textbf{Acc} = \frac{1}{M} \sum_{i=1}^M \mathbf{1}(\hat{y}_i = y_i)
    \]
    where $M$ is the total number of samples, $y_i$ the true label, and $\hat{y}_i$ the predicted label.

    \item \textbf{Precision (Prec)} measures the proportion of correct predictions among all predictions for each class:
    \[
    \textbf{Prec} = \frac{1}{C} \sum_{c=1}^C \frac{TP_c}{TP_c + FP_c}
    \]
    where $C$, $TP_c$, and $FP_c$ are the number of classes, true positives and false positives, respectively for class $c$.

    \item \textbf{Recall (Rec)} measures the proportion of correct predictions among all speech signals for each class:
    \[
    \textbf{Rec} = \frac{1}{C} \sum_{c=1}^C \frac{TP_c}{TP_c + FN_c}
    \]
    where $FN_c$ is the number of false negatives for class $c$.

    \item \textbf{F1-Score (F1)} is the harmonic mean of precision and recall, computed per class and then averaged:
    \[
    \textbf{F1} = \frac{1}{C} \sum_{c=1}^C \frac{2 \cdot \text{Prec}_c \cdot \text{Rec}_c}{\text{Prec}_c + \text{Rec}_c}
    \]

    \item \textbf{Confusion Matrix (CM)} is a $C \times C$ matrix summarizing prediction outcomes; entry $(i,j)$ indicates how many speech signals from class $i$ were predicted as class $j$:
    \[
    \textbf{CM}_{i,j} = \sum_{k=1}^{M} \mathbf{1}(y_k = i \wedge \hat{y}_k = j)
    \]
\end{itemize}

The attribution experiments comprise two scenarios: closed-set and open-set.

\textbf{Experiment 1 - Closed-Set Attribution:} In this scenario, speech signals in the test set belong to the training classes ($y \in \{0,1,\dots,N-1\}$). 

All models output a $(N)$-dimensional softmax vector $p$, where $p_i$ is the predicted probability that $s$ was generated by class $i$. 

The predicted synthesizer class $\hat{y}$ is obtained as:
\[
\hat{y} = \arg\max_i p_i
\]

\textbf{Experiment 2 - Open-Set Attribution:}
In this scenario, speech signals in the test set may originate from either known (seen during training) or unknown (unseen during training) generators ($y \in \{0,1,\dots,N\}$). 

All models output a $(N+1)$-dimensional softmax vector $p$, where $p_i$ is the predicted probability that $s$ was generated by class $i$. $p_N$ represents the predicted probability for the unknown class.

To decide if a speech comes from an unknown generator, we compute the softmax confidence ratio:
\[
r = \frac{\max_i p_i}{\max_{j \neq i} p_j}
\]
A threshold $\delta$ is selected on a 10\% held-out portion of the test set (including known and unknown samples), by maximizing the balanced true positive rates across all classes. If $r > \delta$, we assign $\hat{y} = \arg\max_i p_i$; otherwise, we assign $\hat{y} = N$ (unknown generator).
\section{Detection Results}\label{sec:results}

This section presents the detection results for the experiments described in Section~\ref{sec:experiments}.

\subsection{Experiment 1: Training on ASVspoof2019 (English)}

The results shown in Table~\ref{tab:detection_results1} reveal a  decline in  performance when evaluated on datasets in languages other than English. While the detection methods trained on ASVspoof2019 achieve strong performance on the evaluation set of ASVspoof2019, their EER degrades on datasets in other languages. This highlights the limited generalization capability of these methods to languages other than English. 
Wav2Vec2-AASIST demonstrates relatively better generalization than the other detection methods, as shown in Table~\ref{tab:detection_results1}. 
Although its performance also declines compared to ASVspoof2019, it consistently outperforms other detectors on Spanish datasets. 
This superior performance may stem from several factors, including its transformer-based architecture and extensive self-supervised pre-training, which could enhance cross-lingual capabilities.

Overall, the results of Experiment 1 highlight the importance of developing language-specific datasets, such as Spanish, to effectively train synthetic speech detectors.

\begin{table}[t!]
\caption{EER (in \%) of detectors trained on ASVspoof2019.}
    \label{tab:detection_results1}
    \centering
    \renewcommand{\arraystretch}{1.2}
    \setlength{\tabcolsep}{4.5pt}
    \begin{tabular}{l c c c c}
        \toprule
        \textbf{Method} & $\mathbf{U_{ASV}}$$\downarrow$ & $\mathbf{U_{ODSS}}$$\downarrow$ & $\mathbf{U_{ODSS_{Spa}}}$$\downarrow$ & $\mathbf{U_{HIS}}$$\downarrow$\\
        \midrule
        \textbf{LFCC+GMM} & 3.59\% & 28.74\% & 41.88\% & 42.71\%\\
        \textbf{MFCC-ResNet} & 13.07\% & 49.25\% & 49.57\% & 41.72\%\\
        \textbf{Spec-ResNet}  & 11.04\% & 49.48\% & 48.72 \% & 43.23\% \\
        \textbf{PaSST}  & 4.77\% & 47.98\% &  35.04\% & 32.14\% \\
        \textbf{Wav2Vec2-AASIST} & 0.27\% & 34.08\% & 17.95\% & 19.92\%\\
        \bottomrule
    \end{tabular}
\end{table}

\subsection{Experiment 2: Training on ODSS (Multilingual)}

When detection methods are trained on the multilingual ODSS dataset, the performance distribution across test sets shifts notably compared to Experiment 1 as shown in Table~\ref{tab:detection_results2}. 
In particular, all methods except for LFCC+GMM show clear improvements in EER on the Spanish datasets, such as ODSS Spanish and HISPASpoof, suggesting that multilingual training improves generalization across languages.

\begin{table}[H]
    \caption{EER (in \%) of detectors trained on ODSS.}
    \label{tab:detection_results2}
    \centering
    \renewcommand{\arraystretch}{1.2}
    \setlength{\tabcolsep}{4.5pt}
    \begin{tabular}{l c c c c}
        \toprule
        \textbf{Method} & $\mathbf{U_{ASV}}$$\downarrow$ & $\mathbf{U_{ODSS}}$$\downarrow$ & $\mathbf{U_{ODSS_{Spa}}}$$\downarrow$ & $\mathbf{U_{HIS}}$$\downarrow$\\
        \midrule
        \textbf{LFCC+GMM} & 22.47\% & 10.10\% & 19.66\% & 45.38\%\\
        \textbf{MFCC-ResNet} & 41.32\% & 32.34\% & 33.33\% & 34.44\%\\
        \textbf{Spec-ResNet}  & 39.39\% & 39.67\% & 23.93\% & 32.50\% \\
        \textbf{PaSST}  & 18.55\% & 16.39\% & 6.84\% & 14.05\% \\
        \textbf{Wav2Vec2-AASIST} & 18.79\% & 21.77\% & 15.15\% & 6.73\%\\
        \bottomrule
    \end{tabular}
\end{table}

While this improvement comes with a moderate drop in performance on the evaluation set of ASVspoof2019, the overall results highlight the benefits of multilingual training. 

\subsection{Experiment 3: Training on ODSS Spanish Subset (Spanish)}

In this experiment, the performance across test sets is relatively modest as shown in Table~\ref{tab:detection_results3}. This is particularly evident in terms of generalization, as most methods perform poorly on ASVspoof2019 and ODSS when trained exclusively on Spanish subset of ODSS. In fact, performance remains limited even on HISPASpoof, highlighting the difficulty of training effective detectors with limited language-specific data.

\begin{table}[H]
\caption{EER (in \%) of detectors trained on ODSS Spanish Subset.}
    \label{tab:detection_results3}
    \centering
    \renewcommand{\arraystretch}{1.2}
    \setlength{\tabcolsep}{4.5pt}
    \begin{tabular}{l c c c c}
        \toprule
        \textbf{Method} & $\mathbf{U_{ASV}}$$\downarrow$ & $\mathbf{U_{ODSS}}$$\downarrow$ & $\mathbf{U_{ODSS_{Spa}}}$$\downarrow$ & $\mathbf{U_{HIS}}$$\downarrow$\\
        \midrule
        \textbf{LFCC+GMM} & 22.91\% & 41.69\% & 1.09\% & 45.51\%\\
        \textbf{MFCC-ResNet} & 45.31\% & 46.52\% & 46.15\% & 46.28\%\\
        \textbf{Spec-ResNet}  & 44.84\% & 47.98\% & 37.18\% & 48.94\% \\
        \textbf{PaSST}  & 24.76\% & 14.67\% & 0.85\% & 11.81\% \\
        \textbf{Wav2Vec2-AASIST} & 21.77\% & 38.92\% & 31.62\% & 22.03\%\\
        \bottomrule
    \end{tabular}
\end{table}

The Spanish subset of ODSS suffers from limited size and generator diversity, which likely constrains performance.
When disjoint speaker and generator constraints are applied, the effective number of training speech signals becomes even smaller. This affects all methods, including Wav2Vec2-AASIST, which, unlike previous experiments, fails to maintain a clear performance advantage. This may be due to its higher number of trainable parameters, which makes it more susceptible to underfitting when fine-tuning on limited data. 
Interestingly, we observe that pretrained methods such as PaSST and Wav2Vec2-AASIST, as well as LFCC-GMM, tend to achieve better results under these conditions.


\vspace{0.2cm}
\subsection{Experiment 4: Training on HISPASpoof (Spanish)}

When detection methods are trained on HISPASpoof, we observe significantly improved performance on Spanish speech generated by recent TTS synthesizers, highlighting the benefits of large-scale language-specific training data (see Table~\ref{tab:detection_results4}).
Compared to training on the ODSS Spanish subset (Experiment 3), all methods show substantial improvements when trained on HISPASpoof.

The performance gains are particularly significant for LFCC-GMM and Spec-ResNet, which achieve remarkably low EERs of 1.57\% and 0.72\%, respectively on the HISPASpoof test set.
Even when evaluated on out-of-domain ODSS Spanish Subset, these methods maintain superior performance compared to their corresponding results in Experiment 3.
\begin{table}[b!]
\caption{EER (in \%) of detectors trained on HISPASpoof. Bold values indicate reduced EER as compared to Table~\ref{tab:detection_results3} (Experiment 3).}
    \label{tab:detection_results4}
    \centering
    \renewcommand{\arraystretch}{1.2}
    \setlength{\tabcolsep}{4.5pt}
    \begin{tabular}{l c c c c}
        \toprule
        \textbf{Method} & $\mathbf{U_{ASV}}$$\downarrow$ & $\mathbf{U_{ODSS}}$$\downarrow$ & $\mathbf{U_{ODSS_Spa}}$$\downarrow$ & $\mathbf{U_{HIS}}$$\downarrow$ \\
        \midrule
        \textbf{LFCC+GMM} & 33.64\% & \textbf{14.75}\% & \textbf{0.85}\% & \textbf{1.57}\%\\
        \textbf{MFCC-ResNet} & 47.98\% & 50.07\% & 48.72\% & \textbf{5.17}\%\\
        \textbf{Spec-ResNet}  & \textbf{38.37}\% & \textbf{41.77}\% & \textbf{17.09}\% & \textbf{0.72}\% \\
        \textbf{PaSST}  & 32.97\% & 35.33\% & 17.95\% & \textbf{4.10}\% \\
        \textbf{Wav2Vec2-AASIST} & \textbf{16.28}\% & \textbf{30.54}\% & 43.59\% & \textbf{10.27}\%\\
        \bottomrule
    \end{tabular}
\end{table}
MFCC-ResNet shows limited generalization performance, consistent with the observations in Experiments 2 and 3.


Generalization to other language datasets, especially ASVspoof2019 and multilingual ODSS, remains limited. Most methods experience a substantial increase in EER when evaluated on these cross-lingual test sets.  
Nonetheless, the performance drop is more pronounced when methods are trained on English speech and evaluated on Spanish, compared to the reverse scenario, suggesting asymmetric cross-lingual generalization patterns.

\subsection{Detection Summary}

The results in Tables~\ref{tab:detection_results1},~\ref{tab:detection_results2},~\ref{tab:detection_results3}, and~\ref{tab:detection_results4} highlight 
the importance of language-specific datasets such as HISPASpoof.

Methods trained solely on ASVspoof2019 (English) exhibit consistently poor performance on Spanish datasets, with EER exceeding 40\% for most detectors. This outcome 
confirms that synthetic speech detection remains a language-sensitive task. 
Training on ODSS, a multilingual dataset, leads to improvements in EER.
However, EERs remain high when evaluated on HISPASpoof, indicating that existing multilingual corpora have limited representation of Spanish phonological characteristics.
Methods trained on the Spanish subset of ODSS perform inconsistently: while some detectors benefit from fine-tuning, others suffer from overfitting or underfitting.
Methods trained on HISPASpoof demonstrate  performance gains across all detectors.
These improvements are observed not only on the HISPASpoof test set but also on the Spanish subset of the ODSS dataset.


\section{Attribution Results}

This section presents the attribution results for the experiments described in Section~\ref{sec:attribution-experiments}.

\vspace{-0.3cm}
\subsection{Experiment 1: Closed-set Attribution}

Table~\ref{tab:attribucio_experiment1} presents the results for the closed-set attribution task, where all synthesizers in the test set were also seen during training. All methods perform exceptionally well, with PaSST and Wav2Vec2 achieving near-perfect attribution performance. Spectrogram-based and MFCC-based methods also show strong attribution performance, with scores exceeding 96\% across all metrics.

\begin{table}[H]    
    \centering
    \vspace{-0.1cm}\renewcommand{\arraystretch}{1.15}
    \caption{Results for closed-set attribution using HISPASpoof.}
    \setlength{\tabcolsep}{8pt}
    \begin{tabular}{l c c c c}
        \toprule
        \textbf{Method} & $\mathbf{Acc}$$\uparrow$ & $\mathbf{F1}$$\uparrow$ & $\mathbf{Prec}$$\uparrow$ & $\mathbf{Rec}$$\uparrow$\\
        \midrule
        \textbf{MFCC-ResNet} & 99.11\% & 96.90\% & 98.99\% & 97.88\% \\
        \textbf{Spec-ResNet}  & 99.91\% & 99.87\% & 99.69\% &  99.78\% \\
        \textbf{PaSST}  & 100\% &  100\% & 100\% & 100\%\\
        \textbf{Wav2Vec2-AASIST} & 99.96\% & 99.93\% & 99.86\% & 99.89\% \\
        \bottomrule
    \end{tabular}
    \vspace{-0.1cm}
    \label{tab:attribucio_experiment1}
\end{table}

\subsection{Experiment 2: Open-set Attribution}

Table~\ref{tab:attribucio_experiment2} shows the results for the open-set attribution scenario on HISPASpoof, where XTTS-v1 and Fish-Speech were not included during training. Performance drops across all methods compared to the closed-set scenario, as expected. 
PaSST exhibits the least performance degradation, followed by Spec-ResNet.

\begin{table}[H]
    \centering
    \renewcommand{\arraystretch}{1.1}
    \caption{Results for open-set attribution using HISPASpoof.}
    \setlength{\tabcolsep}{8.5pt}
    \begin{tabular}{l c  c c c}
        \toprule
        \textbf{Method} & $\mathbf{Acc}$$\uparrow$  & $\mathbf{F1}$$\uparrow$ & $\mathbf{Prec}$$\uparrow$ & $\mathbf{Rec}$$\uparrow$\\
        \midrule
        \textbf{MFCC-ResNet} & 43.05\%  & 63.08\% & 80.12\% & 54.29\% \\
        \textbf{Spec-ResNet}  & 69.73\% & 71.22\% & 86.71\% &  71.85\% \\
        \textbf{PaSST}  & 78.32\% &  77.00\% & 91.18\% & 79.67\%\\
        \textbf{Wav2Vec2-AASIST} & 45.57\% & 65.60\% & 83.35\% & 60.28\% \\
        \bottomrule
    \end{tabular}
    \vspace{-0.1cm}
    \label{tab:attribucio_experiment2}
\end{table}

\begin{figure*}[t!]
\vspace{-0.1cm}
    \centering
    \includegraphics[width=0.235\textwidth]{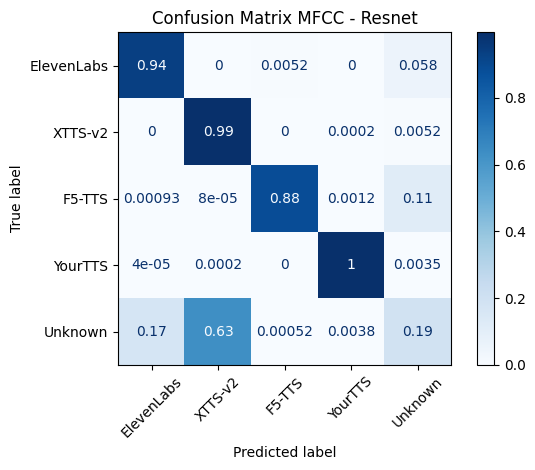}\quad
    \includegraphics[width=0.235\textwidth]{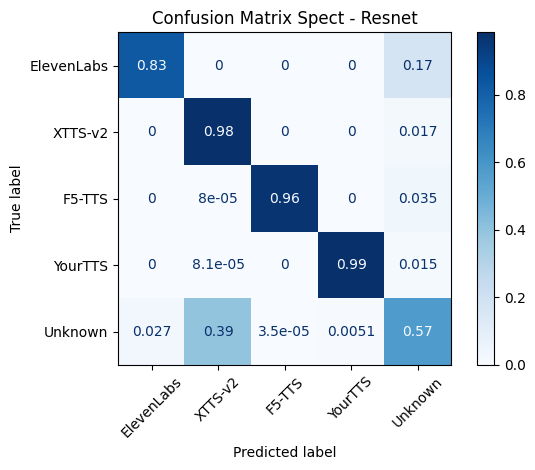}\quad
    \includegraphics[width=0.235\textwidth]{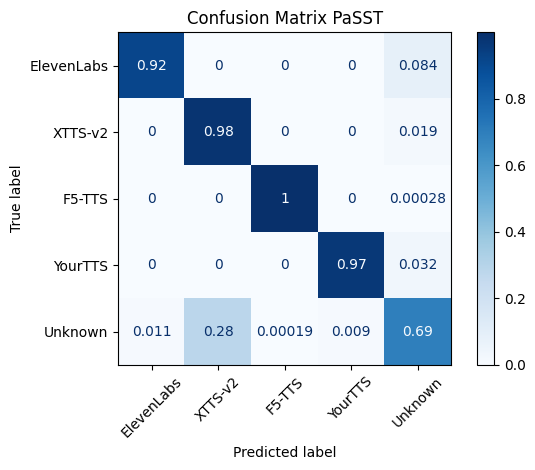}\quad
    \includegraphics[width=0.235\textwidth]{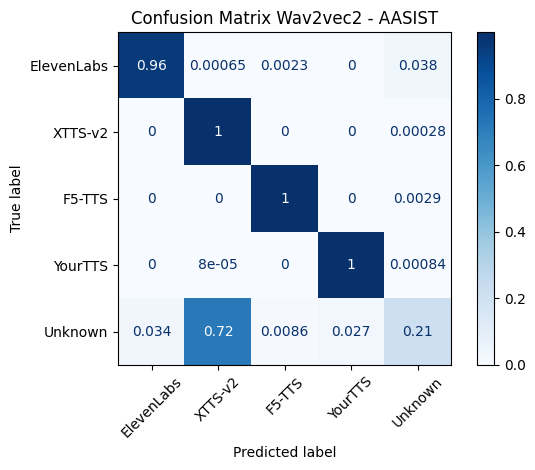}
    \vspace{-0.3cm}
    \caption{Confusion matrices for Attribution Experiment 2: the open-set scenario.}
    \label{fig:cuatro_fotos}
    \vspace{-0.5cm}
\end{figure*}

Fig.~\ref{fig:cuatro_fotos}. shows normalized confusion matrices of all methods. The x-axis represents predicted classes, the y-axis presents true classes, and rows are normalized to form probability distributions summing to one, with minor deviations due to rounding. Correct predictions appear along the diagonal, and unseen generators are classified as ``Unknown". We observe that significant number of unknown speech signals are being misclassified as XTTS-v2.
This pattern is particularly noteworthy given that the unseen generators in this experiment are XTTS-v1 and Fish-Speech. 
The frequent misclassifications of unknown speech signals as XTTS-v2 can likely be attributed to the architectural similarity between the two versions of XTTS, making them difficult to distinguish based on their synthetic speech characteristics.
These results suggest that the methods effectively capture the distinctive characteristics of ElevenLabs, F5-TTS, and Your-TTS, but struggle to differentiate between architecturally similar systems like XTTS-v1 and XTTS-v2. These findings explain the results in Table~\ref{tab:attribucio_experiment2}.



\subsection{Attribution Summary}
Attribution experiments (Tables~\ref{tab:attribucio_experiment1} and \ref{tab:attribucio_experiment2}) confirm the broader utility of the HISPASpoof dataset. In the closed-set scenario, methods achieve near-perfect performance. More importantly, the open-set results reveal that attribution remains a challenging but feasible task: PaSST and Spec-ResNet show better generalizability to unseen generators, paving the way for future work on open-set attribution and modeling the unknown-class.
\glsresetall

\section{Conclusion and Future Work}\label{sec:conclusion}

In this paper, we proposed HISPASpoof, a large-scale Spanish dataset for synthetic speech detection and attribution.
We demonstrated that synthetic speech detectors trained on English generalize poorly to Spanish
and training on HISPASpoof enables better performance on Spanish speech.
We also evaluated synthetic speech attribution performance on HISPASpoof.
Future work will explore training detectors in scenarios where speech data in specific languages is scarce. 
Furthermore, addressing the technical challenges of cross-lingual generalization will require systematic development of multilingual datasets, training approaches and evaluation protocols.

\section*{Acknowledgments}
This material is partially based on research sponsored by DARPA and Air Force Research Laboratory (AFRL) under agreement number FA8750-20-2-1004. The U.S. Government is authorized to reproduce and distribute reprints for Governmental purposes notwithstanding any copyright notation thereon. The views and conclusions contained herein are those of the authors and should not be interpreted as necessarily representing the official policies or endorsements, either expressed or implied, of DARPA and Air Force Research Laboratory (AFRL) or the U.S. Government. Address all correspondence to Edward J. Delp, ace@purdue.edu.

\section*{References}
\setlength{\bibitemsep}{0.3ex}
\AtNextBibliography{\fontsize{7.5}{9}\selectfont}
\printbibliography[heading=none]

\end{document}